# Learning optimal Bayesian prior probabilities from data

Ozan Kaan Kayaalp

**Abstract**

Noninformative uniform priors are staples of Bayesian inference, especially in Bayesian machine learning. This study challenges the assumption that they are optimal and their use in Bayesian inference yields optimal outcomes. Instead of using arbitrary noninformative uniform priors, we propose a machine learning based alternative method, learning optimal priors from data by maximizing a target function of interest. Applying naïve Bayes text classification methodology and a search algorithm developed for this study, our system learned priors from data using the positive predictive value metric as the target function. The task was to find Wikipedia articles that had not (but should have) been categorized under certain Wikipedia categories. We conducted five sets of experiments using separate Wikipedia categories. While the baseline models used the popular Bayes-Laplace priors, the study models learned the optimal priors for each set of experiments separately before using them. The results showed that the study models consistently outperformed the baseline models with a wide margin of statistical significance ($p < 0.001$). The measured performance improvement of the study model over the baseline was as high as 443% with the mean value of 193% over five Wikipedia categories.

## 1 Introduction

The naïve Bayes model is a popular machine learning method due to its simplicity, efficient run-time and high classification performance. Implementers of naïve Bayes and other Bayesian network models usually rely on noninformative uniform priors such as Bayes-Laplace and Jeffreys' priors, widely used by Bayesian statisticians (Jaynes 1968; Good 1983). Assigning meaningful priors is always a challenging task (Petrone et al. 2014). When dealing with Big Data or data on which we have no insight, assigning meaningful priors becomes infeasible; hence, the reliance of the machine learning community on these rather arbitrary, noninformative priors is understandable. However, it is not justifiable to assume that these priors are optimal. In this study, we investigate the questions: Are noninformative uniform priors optimal? If not, can we find optimal priors to improve Bayesian inference for a given target function?

We studied these problems using English Wikipedia as our data source. Authors of Wikipedia categorize their articles with existing (and if necessary, with newly created) Wikipedia categories. Wikipedia offers plenty of already classified data; however, given its size and the nature of Wikipedia, many articles are not categorized with the right categories. In this study, we try to discover the missing article-category links for five Wikipedia categories: (a) *Machine learning*, (b) *Software companies of the United States*, (c) *English-language journals*, (d) *Windows games*, and (e) *American films*. We hope that this study would also encourage the machine learning community to develop new tools to empower Wikipedia communities in their efforts to improving Wikipedia.

## 2 Background

In this section, we review the fundamental aspects of prior probabilities as treated in Bayesian theory, followed by the review of the framework of conditional independence and its use in machine learning as naïve Bayes models. Lastly, we review Wikipedia categories, as they are the classes of interest in our text classification experiments.

### 2.1 Prior probabilities and hyperparameters

The probability of an event $x$ is estimated from a sample usually as the number of observations of that event $n(x)$ divided by the sample size $N$. The implicit assumption is that all events in the sample are independent and identically distributed. If a sample comprises only two types of such events as in the single-coin-toss problem {head, tail}, sample size is the sum of the number of outcomes of heads and tails observed in the sample $N = n(\text{head}) + n(\text{tail})$. The estimated probability of getting the outcome head in a single trial would be $p(\text{head}) = n(\text{head})/N$. This purely frequency based, objective probability estimate is known as the maximum likelihood estimate.

In Bayesian theory, on the other hand, probabilities are estimated not only based on the observation of new events (i.e., data) but also based on personal belief and background knowledge. Since the latter is personal, Bayesian probabilities are qualified as subjective.

When we deal with Big Data, where problems contain a large number of variables, it is very difficult to come up with informative priors to inject the background information into the equation. In such cases, Bayesian practitioners usually follow in the footsteps of Bayes and Laplace, presuming all possible outcomes of an unknown event are equally likely (Jaynes 1968). This approach to priors (a.k.a. Bayes-Laplace priors) implies prior probabilities of binomial events head and tail are equal to 0.5. Since we have no prior information about these events, we should not have any bias towards any of the events, but at the same time, this approach implies that our belief on these events, prior to observing any data, are non-zero.

Another interpretation of priors is that they quantify hypothetically observed historical data or prior equivalent data (PED). If we assume that we had observed two events, one with an outcome of head, $\lambda_H = 1$, and one with tail, $\lambda_T = 1$, our PED size would be equal to 2, the number of events with two PED points. Hyperparameters $\lambda_H$ and $\lambda_T$ are parameters of the prior distributions. In this example, prior probabilities for outcomes head $\lambda_H/(\lambda_H + \lambda_T)$ and tail $\lambda_T/(\lambda_H + \lambda_T)$ are uniform and equal to 1/2. With new observations of size $N = n(\text{head}) + n(\text{tail})$, we would revise our prior beliefs with new likelihoods $n(\text{head})/N$ and $n(\text{tail})/N$:

$$p(\text{head}) = \frac{\lambda_H + n(\text{head})}{(\lambda_H + \lambda_T) + N} \quad \text{and} \quad p(\text{tail}) = \frac{\lambda_T + n(\text{tail})}{(\lambda_H + \lambda_T) + N} \tag{1}$$

Note that the sum of the priors $\lambda_H/(\lambda_H + \lambda_T) + \lambda_T/(\lambda_H + \lambda_T) = 1$ are proper and so are the sum of the revised beliefs, *posteriors*, $p(\text{head}) + p(\text{tail}) = 1$. Furthermore, even if one of the event types (e.g., head) is unobserved (i.e., $n(\text{head}) = 0$), due to non-zero priors, all posteriors would be non-zero as well, i.e. $0 < p(\text{head}) < 1$, so would be their products. Due to this beneficial computational property, Bayes-Laplace priors are sometimes called Laplace smoothing.

In the multinomial case, we estimate probabilities as

$$\frac{\lambda + n_i}{t\lambda + N} \quad \text{where} \quad N = \sum_{i=1}^{t} n_i \tag{2}$$

and $n_i$ is the number of new observations of type *i* out of *t* multinomial types. Here is a summary of the relationships among different noninformative uniform priors using the formula in (2):

- In Bayes-Laplace priors: $\lambda = 1$
- In Jeffreys' priors: $\lambda = 1/2$, and
- In Perks' priors: $\lambda = 1/t$ (Good 1983).

Note that all these priors are uniform (i.e., using the same hyperparameter $\lambda$) across all *t* multinomial types and are proper; i.e., $\sum_t \lambda/t\lambda = 1$.

Bayesian statisticians have sometimes been associated with either traditional or empirical Bayesian School. The traditional approach rejects the idea of determining priors from data (Deely and Lindley 1981), while others do not have such a strict adherence to that philosophy (Casella 1985; Lwin and Maritz 1989). The traditional approach either assigns priors based on belief and prior knowledge, which can be challenging (Petrone et al. 2014), or assigns noninformative uniform priors as described above.

Views and methods on empirical Bayes (EB) are diverse, which make it hard to have a consensus on what constitutes an EB approach (Carlin and Louis 2000). Although their methods may differ, EB approaches converge on the idea of estimating prior distributions from data (Casella 1985; Agresti and Hitchcock 2005).

## 2.2 Bayesian inference with conditional independence assumption

Probability of a particular class $c \in C$ given data $D$ is derived from the Bayes theorem

$$p(c|D) = \frac{p(c)p(D|c)}{p(D)}.$$

The terms in the above formula are stated respectively as

$$posterior = \frac{prior \times likelihood}{evidence}.$$

If the problem has $k$ variables, $D = X_1, \dots, X_k$, and

$$p(c|X_1, \dots, X_k) = \frac{p(c)p(X_1, \dots, X_k|c)}{p(X_1, \dots, X_k)}. \quad (3)$$

An event is a particular instantiations of these variables, $(X_1 = x_1, \dots, X_k = x_k, C = c)$, where $x_i$ is a particular multinomial outcome of variable $X_i$ and $c$ is a particular class in all possible classes of $C$.

In the naïve Bayes model, multinomial variables are conditionally independent given class. The naïve Bayes model computes the probability of a particular event $(x_1, \dots, x_k)$ given the class $c$ as $p(x_1, \dots, x_k|c) = \prod_{i=1}^{k} p(x_i|c)$. Combining this with Equation (3) results Equation (4) denoting the Bayesian inference under the conditional independence assumption, which we used in this study.

$$p(c|x_1, \dots, x_k) = \frac{p(c) \prod_{i=1}^{k} p(x_i|c)}{\sum_{c \in C} \left( p(c) \prod_{i=1}^{k} p(x_i|c) \right)} \quad (4)$$

Although there are a number of variations of this model, all of them are known collectively as naïve Bayes models (McCallum and Nigam 1998).[1]

---

[1] Of the 216,000 Google Scholar articles found with key phrase "text classification", 34,300, roughly 16%, were related to "naïve Bayes".

A common critique against the use of this model for text classification is that the features of the model (words) in reality are not conditionally independent. It is however shown that even when the conditional independence assumption was violated drastically, the classifier usually performs well (Domingos and Pazzani 1997).

## 2.3 Wikipedia article categories

Wikipedia is the largest freely available peer-reviewed written source of general knowledge. The statistic of more than two-billion monthly visits (for English Wikipedia only) indicates its importance for the world. Half of these visits are from the US, where it is the second most popular website after YouTube (Hardwick 2020). Any improvement on Wikipedia's quality would have a multiplier effect to the knowledge created by others (Thompson and Hanley 2018). However, this global treasure has many hidden gems that may not be visible to every user. Many users access Wikipedia either through a link generated by a search engine such as Google, or through the search engine embedded in Wikipedia. Most other users, however, rely on Wikipedia categories.

To see the list of all Wikipedia articles on a certain topic, the user needs to identify the Wikipedia category that is closest to that topic. Tens (if not hundreds) of thousands of research projects rely on Wikipedia and their categories.[2] The success of most of these projects may depend on the quality of the mapping between Wikipedia articles and Wikipedia categories.

As of July 2020, there exist more than 1.4 million Wikipedia categories. For the author of a new Wikipedia article, it is a daunting task to choose the right categories. Inevitably, this process yields suboptimal results and many articles end up associated with categories incompletely. Our study looks into this problem and attempts to answer the following question: Given a set of articles associated with a particular category, how many other articles on the same topic can we discover that were not properly associated with that category?

It is impractical to visit six million Wikipedia pages to find out the ultimate truth, nor would any manual verification by a few individuals yield a reliable result. One approach could be using a random sample of Wikipedia articles to get an estimate. Note, however, the chance of finding a Wikipedia article in any given category is usually less than 0.001, thus a reliable sample containing a reasonable number of articles not properly categorized would still be too large to review manually.

---

[2] A Google Scholar search with keyword "Wikipedia" returns more than 2 million articles. If restricted in article titles (allintitle: Wikipedia), it returns about 19,500 articles.

Given Wikipedia is a voluntary effort of millions of individuals with varying degrees of understanding of the organization of Wikipedia articles, such structural inconsistency is only inevitable. Furthermore, categorizing an article is not an easy task for any Wikipedia author since there are more than 1.4 million categories to choose from. However, the authors are not left alone—Wikipedia editing is a communal effort; thus, the authors receive incredible support from the community. Despite all such support, we are aware of the fact that there are articles that are not associated with proper categories.

## 3 Methods

In this section, we define six orthogonal dimensions of our methodology: (1) model features, (2) model parameters, (3) evaluation metric, (4) cross-validation, (5) hyperparameter space, and (6) search algorithm.

### 3.1 Model features

Traditionally text classification using naïve Bayes model is done using tokenized words. The features can be bag-of-words or set of words. They can be binomial or multinomial. All words from all documents in the training set may be included, or some feature selection may be imposed in order to cap the number of features, hence the size of the model. In our models, we used a set of Boolean features indicating whether the token (a word or a number) occurred in the article, regardless of its frequency.

As seen in equation (4), naïve Bayes is a multiplicative model and conditional probabilities of each feature is independently included into the product. In other words, regardless of how discriminative the feature is, its contribution to the product is yet another value between 0 and 1. The more spurious features (i.e., the more noise) we include, the more the signal washes out.

Although tokens found in positive examples should be representative of the distribution of tokens in the articles to be discovered (or at least, we hope that it is the case), the same cannot be said for the other articles in the Wikipedia since the corpus is huge compared to any reasonable training sample. Since tokens in the negative examples of the training data are poor representations of the distribution of all tokens in Wikipedia, the benefit of their inclusion into the model is questionable at best. Thus, we excluded from our models any tokens that did not occur in the union of positive training cases.

During the classification of a new case, we computed the posterior with the features of the case intersected with the model features. In other words, we did not penalize a case (judged negatively) for the missing tokens that were in the model. Since for each new case, the classification feature set is potentially unique, we call our approach case-specific feature selection.

## 3.2 Model parameterization

Although in section 2.2, we defined how posterior probabilities are computed with priors and likelihoods, the reader (especially the implementers of machine learning systems) may find the following example useful, which comprises a detailed description of the computation of probabilities in Equation (4). Here, we associate the term positive with the category of interest and negative otherwise. In the following example, $\lambda_+$ and $\lambda_-$ denote positive and negative hyperparameters, and $n(x_i, c_+)$ and $n(c_+)$ denote the number of positive articles containing the word $x_i$ and the number of all positive articles, respectively. In this study, we estimated the probabilities for positive cases as follows (for negative cases, simply switch positive signs with negative):

$$p(x_i, c_+) = \frac{\lambda_+ + n(x_i, c_+)}{\lambda_+ + \lambda_- + N}$$

$$p(c_+) = \frac{\lambda_+ + n(c_+)}{\lambda_+ + \lambda_- + N}, \text{and}$$

$$p(x_i|c_+) = \frac{p(x_i, c_+)}{p(c_+)} = \frac{\lambda_+ + n(x_i, c_+)}{\lambda_+ + n(c_+)}$$

## 3.3 Evaluation metric

Given Wikipedia category, the model of interest should discover as many new articles that were supposed to be associated with that category high on the ranked list. The most suitable metric for this task is positive predictive value ($PPV$), which is also known as precision. A positively predicted set of cases comprises 0 or more true positive (TP) and false positive (FP) cases. $PPV$ is the ratio of true positive cases in the set of positively predicted cases.

$$PPV = \frac{n(\text{TP})}{n(\text{TP}) + n(\text{FP})}$$

## 3.4 Cross-validation

Supervised learning is conducted usually using a *k*-fold cross-validation technique where the learning program is executed *k* times using *k* disjoint sets of cases. Each time one disjoint set is used for testing, the remaining $k - 1$ sets are used for training. Since each execution is a hassle, *k* is fixed usually at a small number, such as 5 or 10. Only in rare occasions, especially when the data is very small, the most extreme choice of *k*, *k* equals to the number of all training cases, is adopted. It is called leave-one-out cross-validation. In Bayesian parameter learning, leave-one-out is the best cross-validation choice, not only due to its high performance and most effective use of the training data, but also due to its surprising simplicity in this context.

The technique is quite straightforward. One needs to include all training cases into the initial model which is the union of all cases: $\mathcal{M}_0 = \cup_{i=1}^{k} C_i$. For each validation case $i: 1 \rightarrow k$, the case is excluded from the model, i.e. $\mathcal{M}_i = \mathcal{M}_0 - C_i$. In the coin-toss example, if the validation case is a tail, the parameters in equations (1) are adjusted as

$$p(\text{head}) = \frac{\lambda_H + n(\text{head})}{(\lambda_H + \lambda_T) + N - 1} \quad \text{and} \quad p(\text{tail}) = \frac{\lambda_T + n(\text{tail}) - 1}{(\lambda_H + \lambda_T) + N - 1}$$

### 3.5  Hyperparameter space

The baseline model uses Bayes-Laplace priors; thus, the hyperparameters for positive and negative categories are $\lambda_+ = \lambda_- = 1$. In order to be inclusive, we added into the hyperparameter search space the hyperparameters for the Jeffreys' priors $\lambda_+ = \lambda_- = 1/2$ as well. For any hyperparameter that is greater than 1, we included only natural numbers into the hyperparameter space and capped the space at $\max(\lambda) = 200$. Since we included rational number hyperparameters for Jeffreys' priors, we extended the hyperparameter space in that direction by including $1/10$ and $1/100$ as well.

The search space is not constrained with pairwise-uniform hyperparameters; that is, $\lambda_+$ can be greater than, less than, or equal to $\lambda_-$. However, in this study, we applied the same set of positive and negative priors (i.e., $\lambda_+/(\lambda_+ + \lambda_-)$ and $\lambda_-/(\lambda_+ + \lambda_-)$) for estimating all conditional and marginal probabilities involving positive and negative categories, respectively.

### 3.6  Search algorithm

Since each hyperparameter can be one of 203 values, $\lambda = [0.01, 0.1, 0.5, 1, 2, \ldots, 200]$, the search space corresponds to a two-dimensional $203 \times 203$ matrix bounded by 0.01 at the low-end and 200 at the high-end $\begin{pmatrix} 0.01, 0.01 & \cdots & 0.01, 200 \\ \vdots & \ddots & \vdots \\ 200, 0.01 & \cdots & 200, 200 \end{pmatrix}$.

If it comes to a point where optimality requires a hyperparameter greater than 200, the contrast between the model and baseline priors would serve us sufficiently well to show our point that Bayes-Laplace priors were far from optimal.

Our search algorithm (see RADIAL-GRADIENT-SEARCH) ran starting at one of those matrix cells with coordinates $(x, y)$ corresponding to $(\lambda_-, \lambda_+)$. The parameters passed to RADIAL-GRADIENT-SEARCH were $(x, y, 203, 203)$. After the search was completed, the procedure was repeated a predetermined number of times, starting at different points in the search space to overcome the local-optima problem. We chose the pair of hyperparameters that maximized the *PPV* measure. Since each search is independent, multiple search processes can run on different CPUs in parallel.

The search algorithm covers a 5 × 5 matrix area each cycle. The starting cell is the center of the 5 × 5 matrix. The center cell (*x*, *y*) is evaluated first (see Line 1 in the algorithm below). In the first cycle, the remaining 24 cells are covered. If one of the 24 cells corresponds to the best pair of priors, the center is shifted there (Line 17 followed by Line 6). The search ends, if the *PPV* performance of the pair of priors of the center cell cannot be improved.

---

RADIAL-GRADIENT-SEARCH (*x*, *y*, *xmax*, *ymax*)

1  $ppv \leftarrow$ EVALUATE-PRIORS (*x*,*y*)
2  $best\_cell \leftarrow (ppv, x, y)$
3  $HashTable(x, y) \leftarrow ppv$
4  **while** (TRUE):
5      $improved \leftarrow$ FALSE
6      $(x, y) \leftarrow best\_cell.coord$
7      **for** $i \leftarrow [-2, .., +2]$:
8          **for** $j \leftarrow [-2, .., +2]$:
9              **if** $\neg(i = 0 \land j = 0)$:
10                 $x' \leftarrow x + i$
11                 $y' \leftarrow y + j$
12                 **if** $x' \geq 0 \land x' < xmax \land y' \geq 0 \land y' < ymax \land HashTable(x', y') = \emptyset$:
13                     $ppv \leftarrow$ EVALUATE-PRIORS (*x'*,*y'*)
14                     $HashTable(x', y') \leftarrow ppv$
15                     **if** $ppv > best\_cell.ppv$:
16                         $improved \leftarrow$ TRUE
17                         $best\_cell \leftarrow (ppv, x', y')$
18      **if** $\neg improved$:
19          **return** $best\_cell$

---

For the sake of brevity in the provided RADIAL-GRADIENT-SEARCH algorithm, we excluded further evaluation of a new cell's potential if its *PPV* measure matches that of $best\_cell$ (see Lines 15–17). In our actual implementation however, **if** $ppv = best\_cell.ppv$, we further compared the new cell (*x'*,*y'*) against the $best\_cell$ based on their sensitivity scores and chose the cell the with highest score.

# 4 Preprocessing Wikipedia

The mapping between Wikipedia pages and Wikipedia categories is many-to-many; i.e., a Wikipedia page may be associated with multiple Wikipedia categories and a Wikipedia category may be associated with multiple Wikipedia pages. There are many types of Wikipedia pages such as disambiguation pages, talk pages, category pages, redirect pages, and article pages, but in this study, we are interested only in article pages (Wikipedia:What is an article? 2020) that contain at least 300 bytes of data (roughly 50 words or more). In raw Wikipedia dump *enwiki-20200701-pages-articles-multistream.xml*, the number of articles that fit to this specification was 5,935,124.

During the preprocessing of the raw data, we extracted each article into a unique file whose filename comprises the first 50 characters of the article title. We excluded any metadata from the raw data. We also excluded references at the end of the article and any further article content following the *References* section (such as the section containing category information). In other words, we separated category information from the corpus of articles and stored it separately in other files as metadata for model training and evaluation purposes.

We tokenized the text into words and numbers, stripped out all punctuations around them, and converted all words to lowercase, but did not reduce them into stemmed or lemmatized forms. To keep such a big dataset organized, we partitioned the files into 1,000 directories, each containing roughly equal amounts of data. This approach allowed us to run our codes in a distributed fashion on multiple cores and on multiple machines in parallel.

Any Wikipedia category may contain subcategories. Theoretically, any article in a subcategory should also be a member of its parent category (see section Subcategorization in (Wikipedia:Categorization 2020)) but in practice, we could not trust this for constructing training data. For example, category *Software companies of the United States* contains subcategory *Google*, which contains an article with title *AlphaGo*, which is not a software company and the article was not about the company but about a program developed at that company.

Categories in general should not contain articles that are part of their subcategories unless their subcategories are non-diffusing (see section Non-diffusing subcategories in (Wikipedia:Categorization 2020)). For example, *American Films* category explicitly states: "*all* American films should be included in this category" (Category:American films 2020). *English-language journals* category had only one subcategory, *General law journals*, with 75 articles, of which 47 (63%) were also directly associated with the main category. Given *English-language journals* had only one subcategory, which did not have any further subcategories, it should had been relatively straightforward for the authors of those articles to follow the general rule. But even in this case, the rule of disjoint category-subcategory articles was not followed most of the time. Most other categories have significantly deep subcategory trees.

We stored a list of all articles associated with each category in a separate file, for 1,423,767 categories. Each category file comprised only those articles that are associated with the category directly. Articles associated with a category through a subcategory were not included. We used these category files to determine which articles to include in training. For example, when we learned the model of *English-language journals*, we used all articles in its category file, but not other articles, some of which were indirectly associated with the category through *General law journals*.

# 5 Experiments

In this study, we conducted five sets of experiments, one for each distinct Wikipedia category. We chose the following categories: (a) *Machine learning*, (b) *Software companies of the United States*, (c) *English-language journals*, (d) *Windows games*, and (e) *American films*. Our intention was to cover a wide spectrum of categories—from a relatively small category (*Machine learning*) with 199 articles to the fourth largest category (*American films*) with 52,568 articles. Our choice of categories was partly due to our limited ability and resources to effectively review and verify positively predicted articles. Had we had chosen categories such as *Living people*, or *Association football midfielders* (about midfielder soccer players), we would have had a harder time to judge whether an article suggested by the model did truly belong to that category.

For each experiment, we collected all Wikipedia articles associated with the category of interest and put them in the training data as positive cases. If there were *k* positive cases in the training data, we randomly selected *k* additional cases from the set of all other Wikipedia articles and put them in the training data as negative cases—the chance of randomly selecting a positive case was quite low.

We sampled negative cases using Python's random number generation library. At the beginning of the random number generation, we set the seed so that (a) our readers can replicate our experiments later, and (b) we could repeat the experiments with different sets of negative cases in a controlled fashion. For each category, we ran five experiments with seeds $0, 1, ..., 4$.

Each set of experiments has two branches: (1) baseline and (2) study. For the model of the baseline branch, hyperparameters were predetermined by the Bayes-Laplace priors and likelihood parameters were learned from the data. The baseline model classified all Wikipedia articles, excluding positive cases of the training data.

In the study branch, we learned hyperparameters of the model from data using leave-one-out cross-validation. In each experiment, we repeated the search nine times. At each time, the search started at a different location to minimize the chance of being stuck in a local maxima. The starting points of search were at $(\lambda_-, \lambda_+) = \{(1,1), (1,8), (1,15), (8,1), (8,8), (8,15), (15,1), (15,8), (15,15)\}$. We chose the starting points between 1 and 15, assuming that the hyperparameters we would end up with would probably be between 0.01 and 20. It was expected (and desirable) that multiple search routines would reach to the global maxima and their search paths would coincide. In that case, since they used the same training set (hence, the same likelihood functions), they would produce the same score for the same pair of hyperparameters. At the end of those nine search routines, we should have a semi-complete picture of the terrain of the hyperparameters. Note that since we run these search routines using five different random seeds, we conducted 45 experiments for each category, 225 for all categories.

Given Wikipedia contained 6 million articles and our random sample of negative cases were very small in comparison, experiments using different seeds for random numbers produced a slightly different picture, suggesting a slightly different pair of hyperparameters. To determine the optimal set of priors, we averaged scores for all explored pairs of hyperparameters produced by experiments with different seeds. We classified all Wikipedia articles using learned priors.

We decided that we could effectively review up to 1,000 articles suggested by each model (baseline and study models) for any given category (i.e., if the results were non-overlapping, the total number of articles that we committed to review was 10,000 minus subcategory articles, which are assumed to be positive).

We selected the top 1,000 classifications of both models, stripped them from their scores, merged them into a single set, sorted the resulting list of bare article titles alphabetically, converted the text file into an html page with links to the corresponding Wikipedia articles, and pasted the hyperlinked list into a column of a spreadsheet for easy tallying. We repeated this process five times, once for each category. This way, we blinded our volunteering annotators from any possible information linking the article to the classifier.

## 6 Results

As described in section 4, we excluded all *category articles* from evaluation, since we parameterized the model with them at the first place; however, we used *subcategory articles* in the evaluation. Recall, subcategory articles were supposed to be members of the main category of interest but not part of the category articles. For example, the article *Deep learning* was not in the set of articles of category *Machine learning* but was in its subcategory *Deep learning*.

We reviewed the top 1,000 baseline model predictions and the top 1,000 study model predictions. We labeled those articles as *new articles* if they should have been associated with the category but were associated with neither the category nor any of its subcategories. In short, given category, the three sets of articles shown in Figure 1 were disjoint sets.

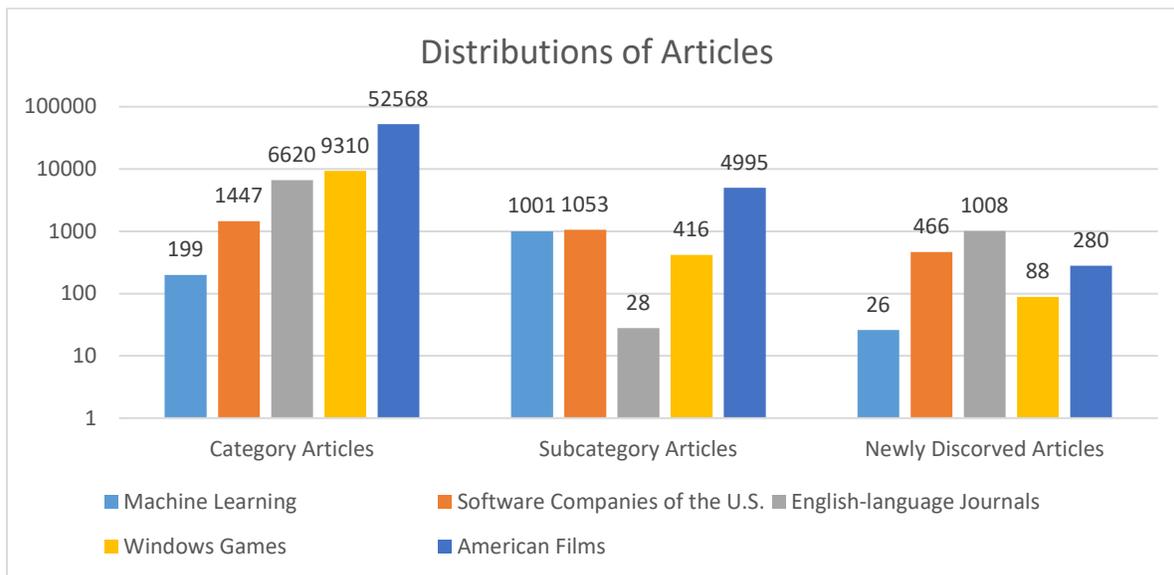

Figure 1 Distributions of category articles, non-overlapping subcategory articles, and newly discovered articles

Note that the category sizes grew exponentially, with the exception of *English-language journals*, whose size was only 29% smaller than the size of category *Windows games*. There was no obvious relation between the category and subcategory sizes, which was another indicator of how different these sets were.

Learning optimal Bayesian priors using the *PPV* metric yielded a distinct set of hyperparameters for these five categories (see Table 1).

Table 1 Optimal Hyperparameters Learned from Data Using *PPV* Metric

|  | $\lambda_-$ | $\lambda_+$ |
|---|---|---|
| Machine Learning | 22 | 4 |
| Software Companies of the U.S. | 35 | 2 |
| English-language Journals | 17 | 1/2 |
| Windows Games | 157 | 1/10 |
| American Films | 200 | 1/2 |

The categories in Table 1 are in ascending order by the category size. The reader can directly observe a relation between the category size (hence, the article count in training) and the prior odds ratio, i.e. $n(article) \propto \lambda_-/\lambda_+$. Note that prior odds ratio is equal to the ratio of the hyperparameters.

This apparent relation however is in fact weak and dubious. First, *English-language journals* category seems an outlier when we look into relation between its *category size* (i.e., the number of category articles in the training data) and the *model size* (i.e., the number of conditionally independent variables in the model) (see Figure 2).

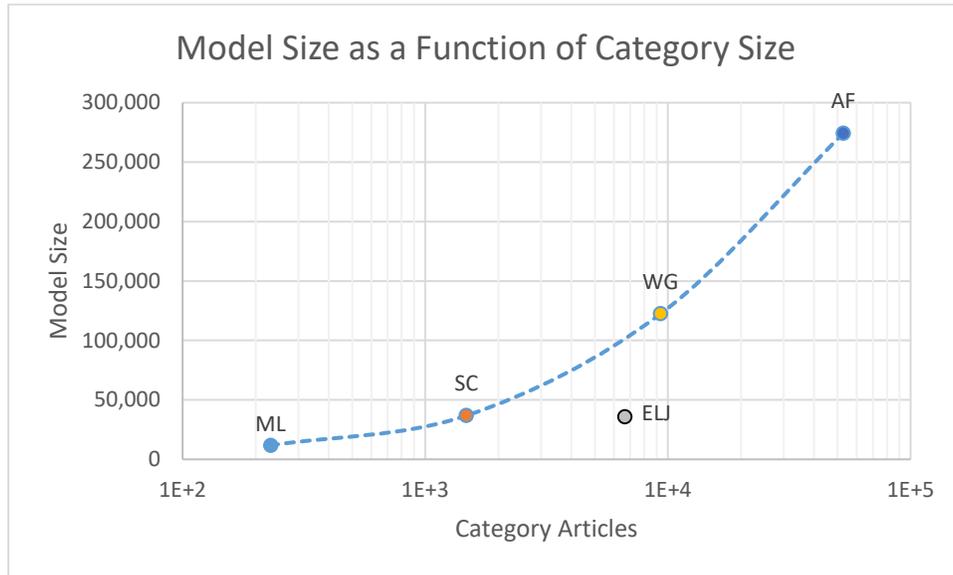

Figure 2 Strong Relation between Category Size and Model Size. ML: *Machine learning*, SC: *Software companies of the US*, ELJ: *English-language journals*, WG: *Windows games*, AF: *American films*.

Second, we also need to take out *American films* from the relation between *model size* and *prior odds ratio* since the search for the hyperparameters of *American films* was cut short at the upper limit of 200. Finally, we are left with only three data points, which are insufficient to establish a strong relation between model size and prior odds ratio (see Figure 3). Further studies and more data points are needed before making any conclusive remark on this issue.

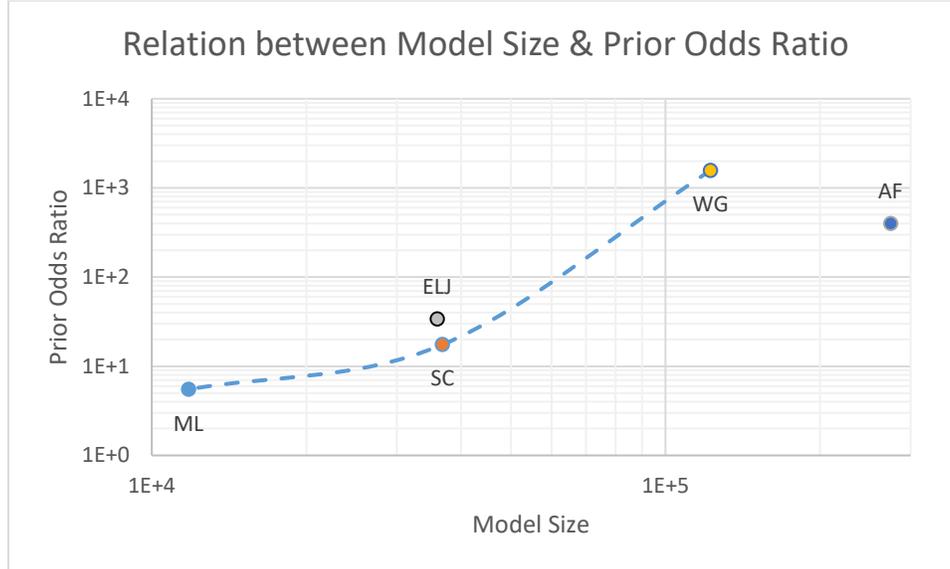

Figure 3 Weak Relation between Model Size and Prior Odds Ratio

A dominating prior lessens the effect of the data on the inference (Box and Tiao 1973). In other words, a high PED raises a high barrier against the features, filtering out swaying effects of the features with low frequencies. In a sense, learned priors function as a feature "selection" system, but unlike most other feature selection approaches, (1) it is integrated into the learning system itself, and more importantly, (2) it is adaptive to the data (as they differed widely among different Wikipedia categories) as well as to the target function of interest (in this study *PPV*).

The baseline and study models made different numbers of positive predictions for each category and the differences were striking (see Figure 4). The study model predicted that 22,239 articles were more likely to belong to the category of *Machine learning* than not; whereas, the baseline models positive predictions were 3.3 million! Both were obviously inaccurate, but the result of the study model was more sensible, not just for machine learning category but for all categories. At the minimum, the positive prediction count of the baseline model was 67% higher than that of the study model, and at the maximum 14,904%. These results indicate that optimal priors calibrated the models significantly.

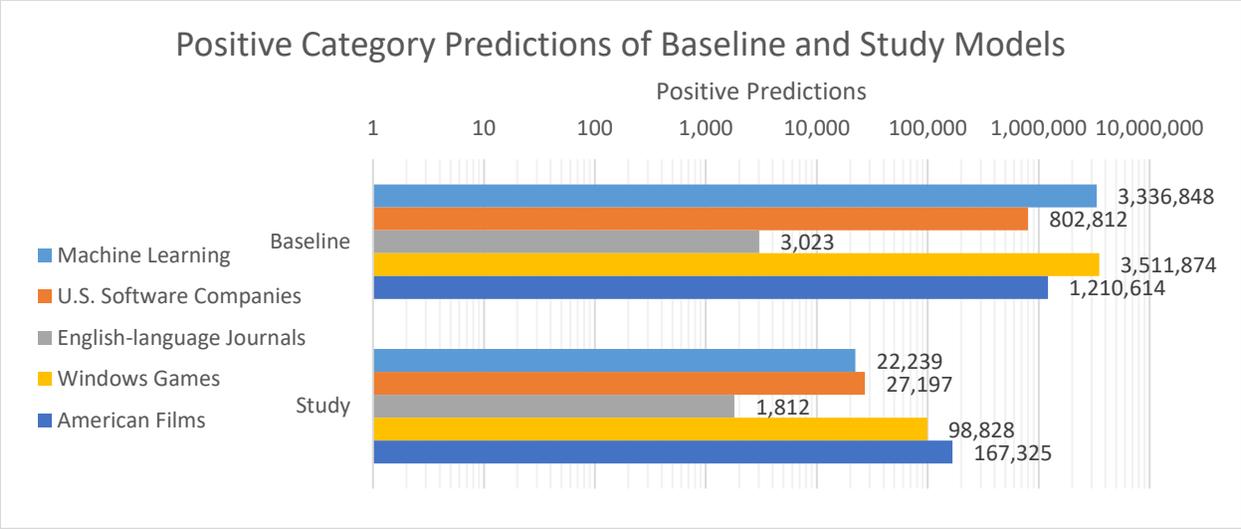

Figure 4 Differences between Baseline and Study Models in Terms of the Number of Positive Predictions

We evaluated the performances of all models by their top-250 positively predicted articles using the *PPV* metric (see Table 2). The improvements of *PPV* measures were meaningfully high with an average rate of 193%. The highest rate was 443% and the lowest 3%.

Table 2 *PPV* Measures of Models Based on their Top 250 Positively Predicted Articles

|  | Baseline | Study | Difference | |
| --- | --- | --- | --- | --- |
| Machine learning | 0.068 | 0.288 | 0.220 | 324% |
| Software companies of the US | 0.092 | 0.500 | 0.408 | 443% |
| English-language journals | 0.928 | 0.960 | 0.032 | 3% |
| Windows games | 0.036 | 0.072 | 0.036 | 100% |
| American films | 0.100 | 0.196 | 0.096 | 96% |

Note that the lowest improvement rate occurred for *English-language journals* when both models scored *PPV* > 0.9. The margin of the improvement at this level could not be as high as in the other cases. For example, even if the study model had a perfect classification result, the improvement by the above analysis would be a mere 8%. A better indicator of improvement at this end of the scale would be how well the study model shrinks the gap between the baseline model's performance and the ideal 1.0. The gap was 0.071, which study model lowered to 0.04, a shrinkage rate of 44%.

Note that low *PPV* scores do not necessarily imply poor classification performance. *PPV* performance is a function of (a) the quality of the classifier, (b) the accuracy of the training data, (c) the quality of text, (d) the number of pertinent articles to be discovered, and (e) granularity/complexity of the classification task. Other than (a), all other factors are the same for both models. The reasons that *English-language* *journals* scored so high may be related to factors (b)–(e). For example, as we see in Figure 2, the model size of the training corpus of *English-language journals* was very small, indicating that article pages were written using shared terms and/or with smaller vocabulary. Therefore, the distribution of their words were much tighter.

Another dimension of the complexity is the clarity and ease of categorization. If the categorization is as straightforward as journal articles, the articles already associated with that category contributes to a solid training set. *Machine learning* category classification was a particularly difficult task, for it was a poorly defined category with no directions for the inclusion and exclusion criteria on its category page. Wikipedia authors included articles not only about different machine learning concepts and methods but also about tasks that machine learning have been used at (e.g., *astrostatistics* and *inauthentic text*). *Machine learning* category also included articles about software products, research labs, concepts used in other fields, data sets, as well as generic problems and puzzles. A further complicating factor was that we were not aware of these facts at the annotation phase and annotated only articles about machine learning concepts and methods as positive. As a result, the ratio of the newly discovered articles over all true positive cases classified by the study model was only 13% for *machine learning* category; whereas it was 95% for *US software compa*nies, 100% for *English-language journals*, 75% for *Windows games*, and 60% for *American films*.

The performance differences between different tasks are also due to the different complexity of the classification tasks. For example, the classification of *English-language journals* basically requires discrimination of journal articles from non-journal articles, since most journals in (English) Wikipedia are in English. *Software companies of the United States* however involves a larger degree of variations. An article may be about a company, which may or may not be a software company. Even if it is a software company, it may not be a US company. Distinguishing articles about a software product from a software company was apparently very challenging as well.

However, the most difficult classification task was for *Windows games* category. There were a number of games developed for smartphones, consoles, and other operating systems that were not ported to Windows platforms. The classification models could not distinguish that level of details since the alternative set (i.e., the set of negative training cases) merely comprised a set of random Wikipedia articles.

To acquire distribution parameters and conduct statistical significance analyses, we resampled the data via non-parametric bootstrap (Efron 1979; Efron and Tibshirani 1986). The results are displayed in Figure 5. In each chart, the line linking the pair of box plots connects the mean values of the baseline and the study. We included the marker × to denote the mean if it did not coincide with the median demarcation line. The whiskers outside of each box denote the first and fourth quartiles—not the confidence intervals. The 95% confidence intervals are in Table 3.

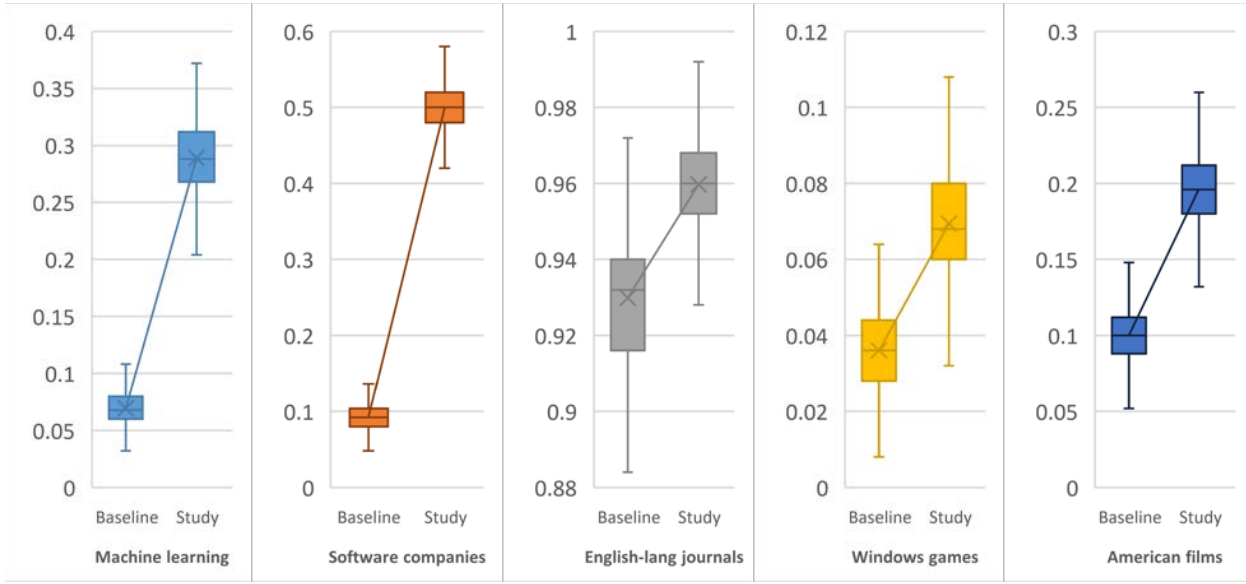

Figure 5 *PPV* Measures of Models Based on their Top 250 Positively Predicted Articles

Table 3 Confidence Intervals (95%) Computed via Bootstrap

|  | Baseline | Study |
|---|---|---|
| Machine Learning | [.040, .104] | [.228, .352] |
| U.S. Software Companies | [.056, .128] | [.432, .564] |
| English-language Journals | [.896, .960] | [.936, .984] |
| Windows Games | [.012, .060] | [.040, .100] |
| American Films | [.064, .140] | [.144, .252] |

We conducted *t*-tests as it was the recommended, most robust statistical significance test for information retrieval (Urbano et al. 2019) and found the performance improvements by the study model was statistically significant ($p < 0.001$) for all pairs of baseline-study model experiments.

Although we compared the performances of the models for the top 250 articles, as reported above we reviewed them for their top 1000 articles. The results of their *PPV* measures displayed in Figures 6-10 were cumulative at the article count on the *x*-axis. For example in Figure 6, the *PPV* measure of the study model at article count 250 was 0.288, corresponding to the number of all true positive cases, which was 72, over 250 positive predictions; i.e., 72/250.

In Figures 6-10, we highlighted the area between the two *PPV* lines with a shade of the line at the top (orange or blue) so that the cross-over points and the dominating model can be recognized easily.

The study model's classification performance was quite stable for machine learning (see Figure 6). The baseline model, however, started with some wrong predictions at first but at around the tenth prediction, it peaked to 0.33 *PPV* and tailed down smoothly. It went below 0.1 at article 121 and ended up with *PPV* = 0.034 at article 1000. The study model followed mostly the same pattern as the baseline model but with higher *PPV*s. It ended up with *PPV* = 0.14. The mean *PPV* difference between two series was 0.17.

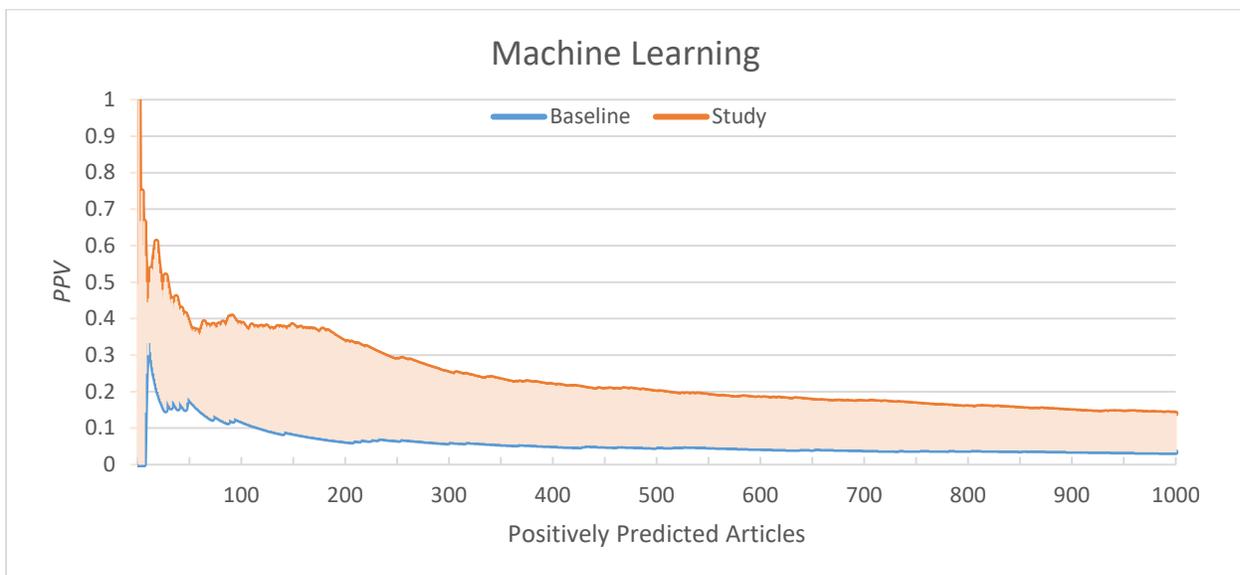

Figure 6 Top 1000 Articles Predicted in Category *Machine Learning*

The article classification performance for category *Software companies of the US* was somewhat similar to that for category machine learning (see Figure 7) although with much higher performance difference between the study and baseline models. The mean *PPV* difference between the two series was 0.38.

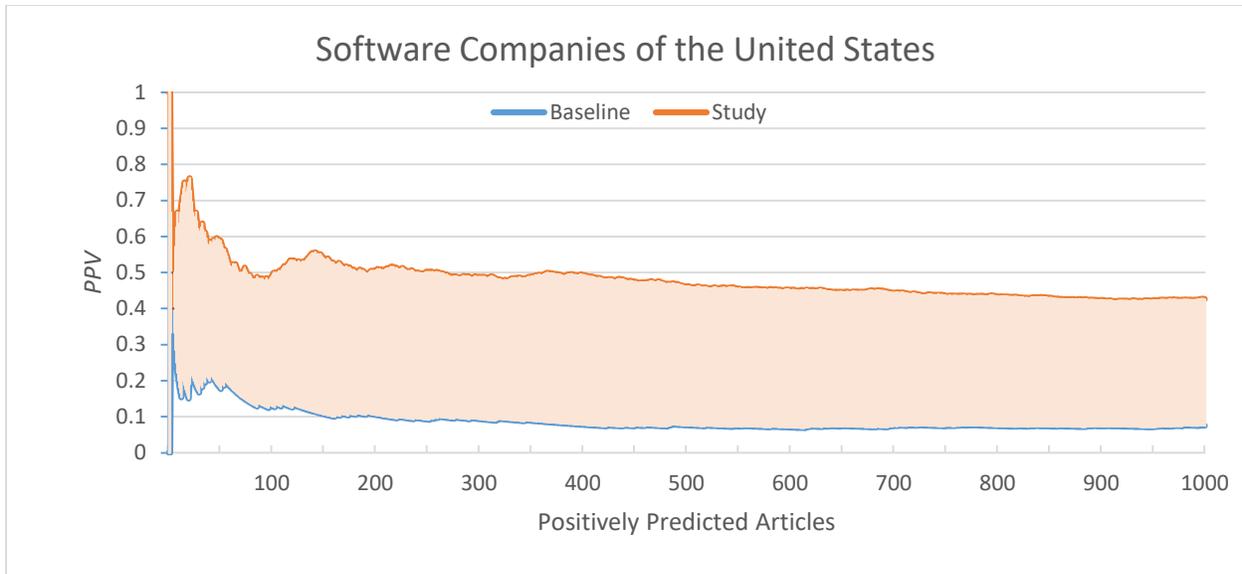

Figure 7 Top 1000 Articles Predicted in Category *Software Companies of the United States*

The *PPV* performance profiles of models for category *English-language journals* in Figure 8 was quite different from the first two. For one, the *PPV* ranges were not as extensive. In Figures 6 and 7, *PPV* measures were displayed on the full range of vertical axis between 0 and 1; whereas, in the series of *English-language journals* the range was between 0.9 and 1. The performance difference between the two series was quite slim with the mean value of 0.014. Yet the study model dominated the baseline model throughout as indicated by the light orange color of the area between these two series. The only exceptions were for predictions between articles 622 and 627. At both points the *PPV* values of both models were identical and between those two points the baseline model performed slightly higher ($\Delta PPV = 0.0016$).

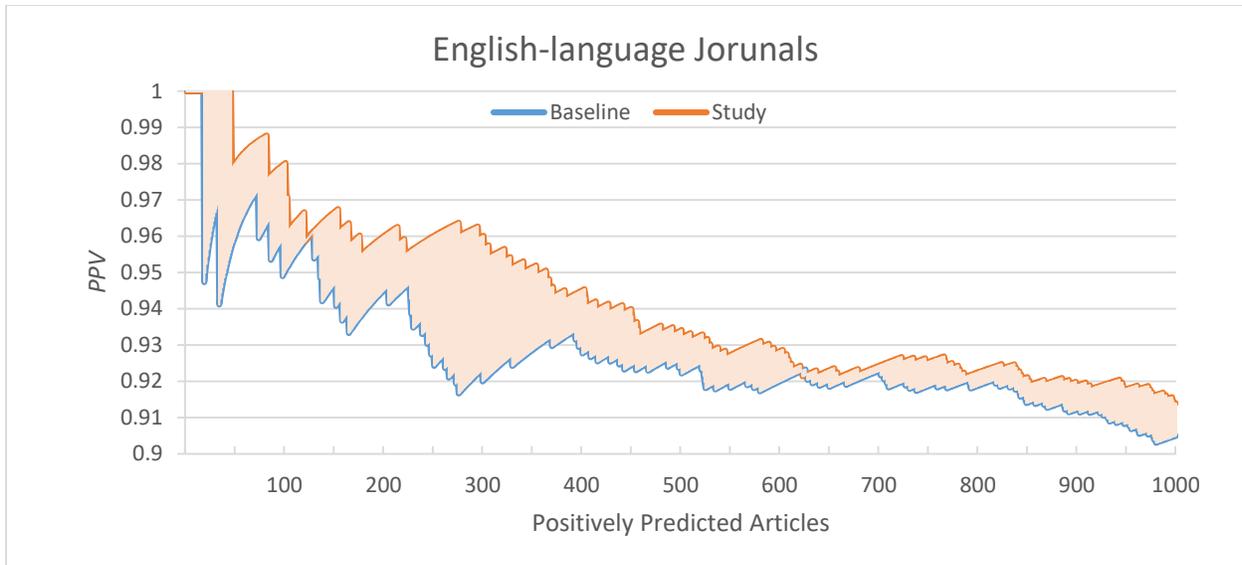

Figure 8 Top 1000 Articles Predicted in Category *English-language Journals*

Please note that the steep decline observed in Figure 8 is actually not steeper than the previous series, but appears to be so due to focus of the display on the top 10% of the *PPV* range, which we trust is obvious to the careful reader.

In the series of *Windows games* classifications (see Figure 9), there was surprisingly no apparent decline but a steady *PPV* performance, albeit with a quite low trajectory, by both models. The fluctuations were higher in the first 200 predictions compared to the last 800, due to averaging over larger numbers. Note however, the mean *PPV* of Windows Games over 1000 predictions were the lowest among all five categories: 0.04 and 0.07 for the baseline and study models, respectively.

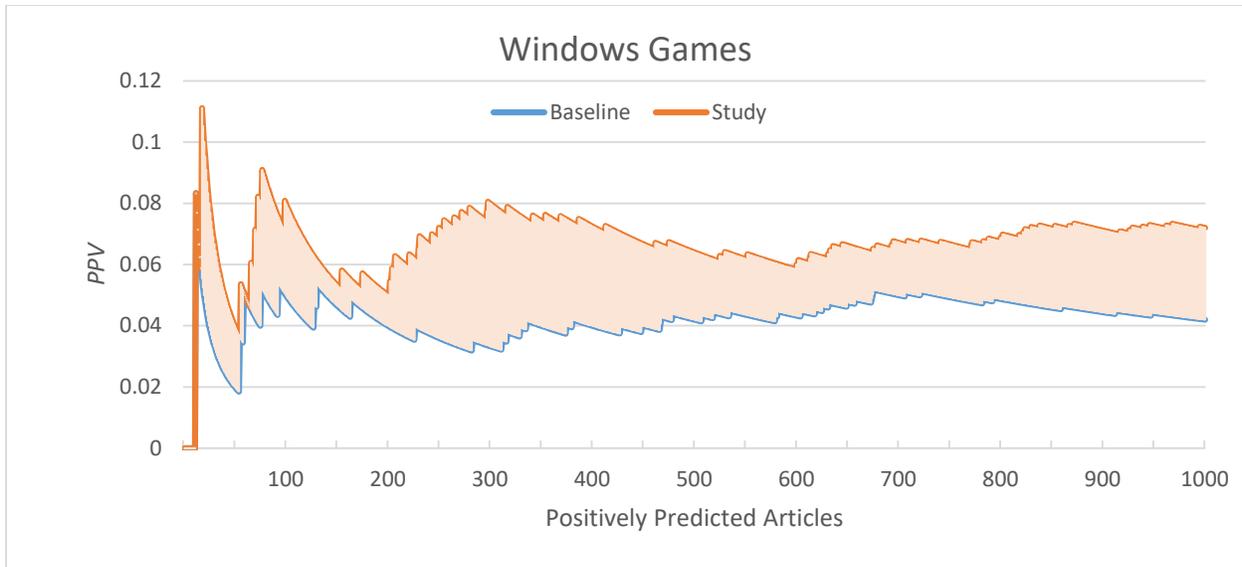

Figure 9 Top 1000 Articles Predicted in Category Windows Games

On *American films* category (see Figure 10), both baseline and study models started with very low scores but steadily improved over the course of the first 1000 predictions. The performance trajectories were quite different from what we observed on the other models. Up until prediction number 98, the race between the baseline and study models was inconclusive as indicated by alternating orange and blue areas between the two model lines. Thereafter, the study line consistently performed better as the distance between the two lines got wider over the course of classifications.

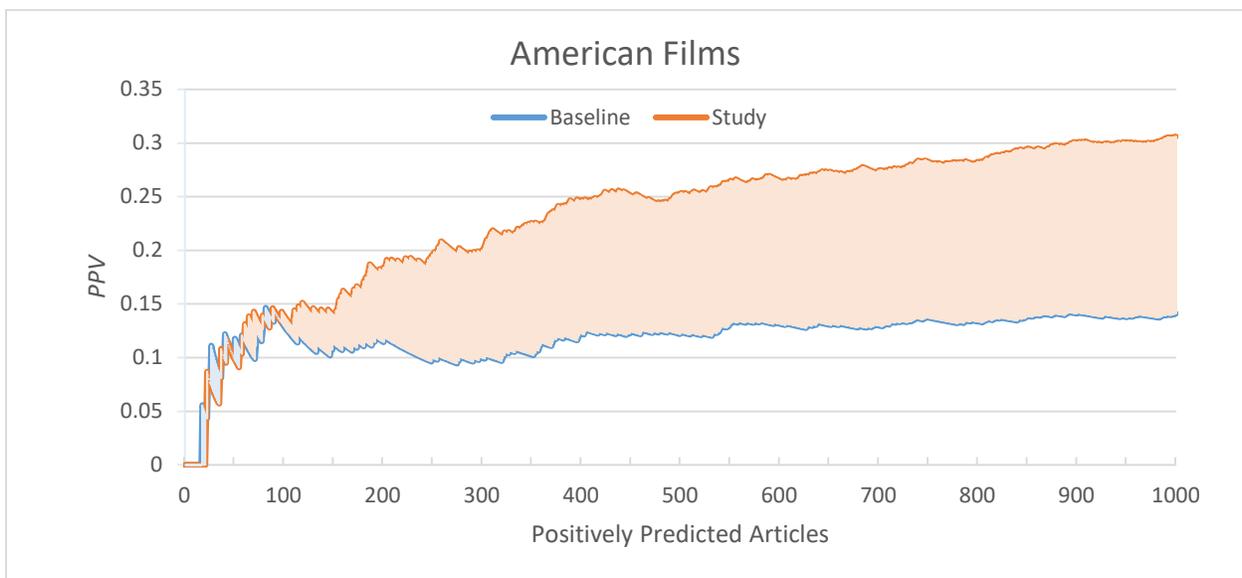

Figure 10 Top 1000 Articles Predicted in Category American Films

# 7 Conclusions

We conducted 235 experiments—225 for hyperparameter search, 10 for evaluations of two models on five categories—to discover Wikipedia articles that should have been associated with certain categories. To evaluate the effectiveness of the study method under different conditions, we carefully chose five distinct categories with varying set sizes of article associations and varying degrees of text classification complexity.

We analyzed the results comparatively based on the top 250 positively classified articles and showed that the performance improvements that the learned optimal Bayesian priors brought were not only statistically significant but also meaningful. Since such a cross-sectional analysis gives only a glimpse of the model behavior at a particular point, we provided detailed *PPV* performance profiles of both models for their 1,000 positive predictions for every category. These performance profiles showed that how drastically these five tasks differed from each other, yet the study models using learned optimal Bayesian priors consistently outperformed the baseline models, which used standard Bayes-Laplace priors.

Given Wikipedia's development is the collaborative effort of millions of individuals, article markups and categorizations are messy. Although we preprocessed all the pages in Wikipedia to the best of our abilities, we are aware of the fact that the preprocessed data is still quite noisy and requires further preprocessing. However, we believe that imperfections in the input had no significant bearing on our experiments, on the results, and on our analyses.

The results showed that the new Bayesian learning method that we introduced in this study can learn optimal prior probabilities from data and can significantly improve the performance of Bayesian inference. We described how efficiently the process of learning optimal priors can be conducted through leave-one-out cross-validation. We described the process on the conditional independence framework of the naïve Bayes model, but the method can be used on any Bayesian network.

To the best of our knowledge, this is the first machine learning study that directly learns prior probabilities from data. There are a number of articles in the literature studying Bayesian methods for learning "hyperparameters of neural network models," which should not be confused with our method for learning "hyperparameters of prior probability distributions of Bayesian models." The term hyperparameters in machine learning studies usually refers to controls of the learning process, not to prior probabilities or prior equivalent data size.

While EB approaches in general aim at maximizing the probability of the observed data, our method has the flexibility of using any target function of choice—due to the current context of efficiently finding certain Wikipedia articles, the target function in this study was *PPV*.

From a vantage point outside of Bayesian statistics, learning priors from data might be viewed as a feature "selection" system. But unlike other feature selection approaches, (1) this method is integrated to the learning system itself, and more importantly, (2) the feature selection thresholds are adaptive to the data and to the target function of interest. However, one may duly argue against this interpretation, since this method does not include or exclude features to/from the model, changing the size of the model; rather, it imposes positive or negative biases on the estimation of their probabilities during model parameterization, making it harder (or easier) for the feature to contribute to the positive or negative classification. In other words, our method acts more like a hindrance (or promoter) to the effect of the feature on a particular classification decision by raising (or lowering) the barriers of priors.

As explained in section 3.2, the standard naïve Bayes model directly uses the likelihood statistic to estimate model parameters. There is no search involved; hence, its learning component is rather rudimentary and debatable. The application of the method of this study transforms the standard naïve Bayes model to a truly learning model (Mitchell 1997) as the algorithm conducts a search through a hyperparameter space to maximize the target function of interest. The results show that the study method improved the performance of the standard model significantly.

We applied our approach to the binomial text classification problem. Extending it to any multinomial Bayesian classification problem is straightforward. As seen in our results, positive and negative class hyperparameters (i.e., $\lambda_-$ and $\lambda_+$) may differ, which we applied uniformly to the parameterization of every feature in the model. In other words, our approach of learning hyperparameters was class-specific. Extending our approach to learning both class- and feature-specific hyperparameters (i.e., $\lambda_{-,x_i}$ and $\lambda_{+,x_i}$) is computationally nontrivial and needs further studies.

## References

Agresti, A., & Hitchcock, D. B. (2005). Bayesian inference for categorical data analysis. *Statistical Methods and Applications, 14*(3), 297-330.
Box, G. E., & Tiao, G. C. (1973). *Bayesian inference in statistical analysis*: Addison-Wesley.
Carlin, B. P., & Louis, T. A. (2000). Empirical Bayes: Past, present and future. *Journal of the American Statistical Association, 95*(452), 1286-1289.
Casella, G. (1985). An introduction to empirical Bayes data analysis. *The American Statistician, 39*(2), 83-87.
Category:American films (2020). https://en.wikipedia.org/w/index.php?title=Category:American_films&oldid=973474161. Accessed December 14 2020.
Deely, J., & Lindley, D. (1981). Bayes empirical bayes. *Journal of the American Statistical Association, 76*(376), 833-841.
Domingos, P., & Pazzani, M. (1997). On the optimality of the simple Bayesian classifier under zero-one loss. *Machine learning, 29*(2-3), 103-130.

Efron, B. (1979). Bootstrap Methods: Another Look at the Jackknife. *The Annals of Statistics, 7*(1), 1-26.

Efron, B., & Tibshirani, R. (1986). Bootstrap methods for standard errors, confidence intervals, and other measures of statistical accuracy. *Statistical science*, 54-75.

Good, I. J. (1983). *Good thinking: The foundations of probability and its applications*: U of Minnesota Press.

Hardwick, J. (2020). Top 100 Most Visited Websites by Search Traffic (as of 2020). https://ahrefs.com/blog/most-visited-websites/. Accessed December 19 2020.

Jaynes, E. T. (1968). Prior probabilities. *IEEE Transactions on systems science and cybernetics, 4*(3), 227-241.

Lwin, T., & Maritz, J. (1989). Empirical Bayes approach to multiparameter estimation: with special reference to multinomial distribution. *Annals of the Institute of Statistical Mathematics, 41*(1), 81-99.

McCallum, A., & Nigam, K. A comparison of event models for naive bayes text classification. In *AAAI-98 workshop on learning for text categorization, 1998* (Vol. 752, pp. 41-48, Vol. 1): Citeseer

Mitchell, T. (1997). Introduction to machine learning. *Machine learning, 7*, 2-5.

Petrone, S., Rizzelli, S., Rousseau, J., & Scricciolo, C. (2014). Empirical Bayes methods in classical and Bayesian inference. *Metron, 72*(2), 201-215.

Thompson, N., & Hanley, D. (2018). Science is shaped by wikipedia: Evidence from a randomized control trial.

Urbano, J., Lima, H., & Hanjalic, A. Statistical significance testing in information retrieval: an empirical analysis of type I, type II and type III errors. In *Proceedings of the 42nd International ACM SIGIR Conference on Research and Development in Information Retrieval, 2019* (pp. 505-514)

Wikipedia:Categorization (2020). https://en.wikipedia.org/w/index.php?title=Wikipedia:Categorization&oldid=990003064. Accessed December 14 2020.

Wikipedia:What is an article? (2020). https://en.wikipedia.org/w/index.php?title=Wikipedia:What_is_an_article%3F&oldid=984465144. Accessed December 14 2020.